\pdfoutput=1
\newif\ifdraft\draftfalse

\newif\ifinlinerefs\inlinerefsfalse
\inlinerefstrue

\documentclass{new_tlp}
\usepackage[T1]{fontenc}
\usepackage{mathptmx}
\usepackage{amsmath,amsfonts,amssymb}
\usepackage{xspace}
\usepackage{color}
\usepackage{booktabs}
\usepackage{pgfplots}\usetikzlibrary{plotmarks}
\usepackage{caption}
\usepackage{subcaption}
\pgfplotsset{
	filter discard warning=false %
	, legend cell align=left
	, minor grid style={loosely dotted, lightgray}
	, major grid style={loosely dashed, lightgray}
}

\ifdraft
\usepackage{showlabels}
\fi

\newcommand\bcmdtab{\noindent\bgroup\tabcolsep=0pt%
  \begin{tabular}{@{}p{10pc}@{}p{20pc}@{}}}
\newcommand\ecmdtab{\end{tabular}\egroup}

\newcommand\lar{\ensuremath{\,\leftarrow\,}}
\newcommand\mi[1]{\ensuremath{\mathit{#1}}}
\newcommand\wasp{\textsc{wasp}\xspace}
\newcommand\clasp{\textsc{clasp}\xspace}
\newcommand\clingo{\textsc{clingo}\xspace}
\newcommand\gringo{\textsc{gringo}\xspace}
\ifdraft
\newcommand\todo[1]{{{\bf TODO:} \textcolor{red}{#1}}}
\else
\newcommand\todo[1]{}
\fi
\newcommand\lazy{\textsc{lazy}\xspace}
\newcommand\eager{\textsc{eager}\xspace}
\newcommand\post{\textsc{post}\xspace}

\newcommand\posOf[1]{\ensuremath{#1_{|_+}}}

\newcommand{\GP}[1]{\ensuremath{Ground(#1)}\xspace}
\newcommand{\HB}[1]{\ensuremath{B_{#1}}\xspace}
\newcommand{\HU}[1]{\ensuremath{U_{#1}}\xspace}

\newtheorem{example}{Example}
\def\naf{\ensuremath{\raise.17ex\hbox{\ensuremath{\scriptstyle\mathtt{\sim}}}}\xspace}
\usepackage[
ruled
, vlined
, linesnumbered]{algorithm2e}

  \title[Constraints, Lazy Constraints, or Propagators]
        {Constraints, Lazy Constraints, or Propagators \\ in ASP Solving: An Empirical Analysis}

  \author[Cuteri et al.]
         {\!\!BERNARDO CUTERI$^1$,\! CARMINE DODARO$^2$,\! FRANCESCO RICCA\!$^1$,\! PETER SCH\"ULLER$^{3,4}$\!\!\\
         $^1$DeMaCS, University of Calabria, Italy\\
         $^2$DIBRIS, University of Genova, Italy\\
         $^3$Faculty of Engineering, Marmara University, Turkey\\
         $^4$Institute of Information Systems, Knowledge-based Systems Group, TU Wien, Austria \\
         \email{\{cuteri,ricca\}@mat.unical.it, dodaro@dibris.unige.it, schueller.p@gmail.com}\\
     }

\newcommand{\portfoliotable}[1]{%
\begin{table}[#1]%
	\caption{Applicability of a portfolio approach: experimental results.} \label{tab:portfolio}%
	\footnotesize%
	\centering%
	\setlength{\tabcolsep}{0.2em}%
	\begin{tabular}{llcccccccc}%
		\multicolumn{2}{c}{} & \multicolumn{4}{c}{\wasp-based} & \multicolumn{4}{c}{\clingo-based}\\
		\cline{3-6} \cline{7-10}\\
		\textbf{Problem} & \textbf{\#instances} & \textbf{Prec} & \textbf{Recall} & \textbf{F-Meas}  & \textbf{Perf. gain}  & \textbf{Prec} & \textbf{Recall} & \textbf{F-Meas}  & \textbf{Perf. gain} \\
		Stable Marriage	&	500	&	0.66	&	0.67	&	0.63	& 26.7\% &	0.74	&	0.75	&	0.74	& 13.6\%\\
		NLU 	& 	150	&	0.88	&	0.92	&	0.90 & 38.3\%&	0.84	&	0.84	&	0.84	& 10.0\%\\
		\bottomrule%
	\end{tabular}%
\end{table}%
}

\begin{document}

\label{firstpage}

\maketitle

  \begin{abstract}
  Answer Set Programming (ASP) is a well-established declarative paradigm.
  One of the successes of ASP is the availability of efficient systems.
  State-of-the-art systems are based on the ground+solve approach.
  In some applications this approach is infeasible because the grounding of one or few constraints is expensive.
  In this paper, we systematically compare alternative strategies to avoid the instantiation of problematic constraints, that are based on custom extensions of the solver.
  Results on real and synthetic benchmarks highlight some
  strengths and weaknesses of the different strategies. \\
  (Under consideration for acceptance in TPLP,
  ICLP 2017 Special Issue.)
  \end{abstract}

\section{Introduction}
Answer set programming (ASP) is a declarative formalism for knowledge representation and reasoning based on stable model semantics \cite{DBLP:journals/ngc/GelfondL91,DBLP:journals/cacm/BrewkaET11}, for which robust and efficient implementations are available~\cite{DBLP:conf/iclp/GebserKKOSW16}.
State-of-the-art ASP systems are usually based on the ``ground+solve'' approach~\cite{DBLP:journals/aim/KaufmannLPS16}, in which a \textit{grounder} module transforms the input program (containing variables) in an equivalent variable-free one, whose stable models are subsequently computed by the \textit{solver} module.
ASP implementations adopting this traditional approach are known to be effective for solving complex problems arising from academic and industrial applications, including: product configuration~\cite{DBLP:conf/scm/KojoMS03}, decision support systems for space shuttle flight controllers~\cite{DBLP:conf/asp/NogueiraBGWB01}, explanation of biomedical queries~\cite{DBLP:journals/tplp/ErdemO15}, construction of phylogenetic supertrees~\cite{DBLP:journals/tplp/KoponenOJS15}, data-integration~\cite{DBLP:journals/tplp/MannaRT15}, reconfiguration systems~\cite{DBLP:conf/cpaior/AschingerDFGJRT11}, and more.
Nonetheless, there are some classes of programs
(cf.\ \cite{DBLP:journals/ai/CalimeriGMR16})
whose evaluation is not feasible with the ``ground+solve'' approach just because the grounding phase induces a combinatorial blow-up.
An issue that is usually referred to as the \textit{grounding bottleneck} of ASP.

The grounding bottleneck has been subject of several studies in recent years, and various alternative approaches to overcome it have been proposed.
Some of these are based on syntactic extensions that enable the combination of ASP solvers with solvers for external theories~\cite{DBLP:journals/tplp/OstrowskiS12,DBLP:journals/corr/BalducciniL17,DBLP:journals/tplp/BalducciniL13,DBLP:journals/tplp/AzizCS13,DBLP:journals/jair/CatDBS15,DBLP:conf/iclp/SusmanL16,DBLP:conf/birthday/EiterRS16};
whereas, the most prominent approach working on plain ASP is \textit{lazy grounding}, which was implemented by \textsc{asperix}~\cite{DBLP:conf/lpnmr/LefevreN09a}, \textsc{gasp}~\cite{DBLP:journals/fuin/PaluDPR09}, and \textsc{omiga}~\cite{DBLP:conf/jelia/Dao-TranEFWW12}.
Roughly, the idea of lazy grounding is to instantiate rules only when it is required during the search for a stable model~\cite{DBLP:journals/ai/LiuPST10}. %
In this way, it is possible to prevent the grounding of rules that are unnecessary for the computation.
Albeit lazy grounding techniques obtained promising preliminary results,
they cannot yet reach the performance of state of the art systems in many
benchmarks~\cite{DBLP:journals/ai/CalimeriGMR16,DBLP:conf/lpnmr/LefevreN09a}.
One of the reasons is probably that fully-fledged lazy grounding techniques could not be easily integrated within solvers based on the very efficient Conflice-Driven Clause Learning (CDCL) algorithm~\cite{DBLP:journals/tc/Marques-SilvaS99,DBLP:journals/aim/KaufmannLPS16,Weinzierl2017}.
Nonetheless, in many applications, the grounding bottleneck is merely caused by rules of a specific kind, namely constraints.
For example, the following constraint has been identified
as the bottleneck in programs solving a problem of natural language understanding:
\begin{equation*}
\leftarrow eq(X,Y), \ eq(Y, Z),\ \naf eq(X, Z)
\end{equation*}
Its grounding, which features a cubic number of instances with respect to the extension of predicate $eq$ in the worst case, is often not feasible for real world instances~\cite{DBLP:journals/fuin/Schuller16}.

In this paper, we focus on the above practically-relevant case of problematic constraints.
In particular, we systematically compare alternative strategies that avoid the instantiation of some constraints by extending a CDCL-based ASP solver.
In a nutshell, the input program is simplified by omitting problematic constraints and it is grounded; then, the resulting ground program is provided as input to a solver that is extended to emulate the presence of missing constraints.
Among the strategies for extending the solver, we considered \textit{lazy instantiation of constraints} and \textit{custom propagators}.
In the first strategy, the solver searches for a stable model $S$ of the simplified program. Then, $S$ is returned as a solution if it satisfies also the omitted constraints, otherwise the violated instances of these constraints are lazily instantiated, and the search continues (Sec.~\ref{sec:lazy}).
In the second strategy, the solver is extended (in possibly alternative ways) by custom \textit{propagators}, which emulate the presence of missing constraints during the search (Sec.~\ref{sec:propagators}).
The above-mentioned strategies can be implemented by using the API of existing CDCL-based ASP solvers~\cite{DBLP:conf/iclp/GebserKKOSW16,DBLP:conf/aiia/DodaroRS16}.
An empirical evaluation conducted on real and synthetic benchmarks (Sec.~\ref{sec:experiments}) confirms that the usage of lazy instantiation and custom propagators is effective when the grounding bottleneck is due to some constraint. The analysis of the results highlights strengths and weaknesses of the different strategies.
Moreover, it shows there is not always a clear winner for a given problem, and the choice depends also on the characteristics the instances to solve.
This observation suggested to investigate the applicability of algorithm selection techniques.
The results are positive, in the sense that already a basic portfolio is faster than the best approach.

\section{Answer Set Programming (ASP)}
An ASP program $\Pi$ is a finite set of rules of the form:
\begin{equation}\label{eq:rule}
  a_1 \lor \ldots \lor a_n \lar b_1, \ldots, b_j, \naf b_{j+1}, \ldots, \naf b_m
\end{equation}
where $a_1,\ldots,a_n,b_1,\ldots,b_m$ are atoms and $n\geq 0,$
$m\geq j\geq 0$.
In particular, an \emph{atom} is an expression of the form $p(t_1, \ldots, t_k)$, where $p$ is a predicate symbol and $t_1, \ldots, t_k$ are \emph{terms}.
Terms are alphanumeric strings, and are distinguished in variables and constants.
According to the Prolog's convention, only variables start with an uppercase letter.
A \emph{literal} is an atom $a_i$ (positive) or its negation $\naf a_i$ (negative), where $\naf$ denotes the \emph{negation as failure}.
Given a rule $r$ of the form (\ref{eq:rule}), the disjunction $a_1 \lor \ldots \lor a_n$ is the {\em head} of $r$, while $b_1,\ldots,b_j, \naf b_{j+1}, \ldots, \naf b_m$ is the {\em body} of $r$, of which $b_1,\ldots,b_j$ is the {\em positive body}, and $\naf b_{j+1}, \ldots, \naf b_m$ is the {\em negative body} of $r$.
A rule $r$ of the form (\ref{eq:rule}) is called a \textit{fact} if $m=0$ and a \textit{constraint} if $n=0$.
An object (atom, rule, etc.) is called {\em ground} or {\em
propositional}, if it contains no variables.
Rules and programs are \textit{positive} if they contain no negative literals, and \textit{general} otherwise. %
Given a program $\Pi$, let the \emph{Herbrand Universe} \HU{\Pi} be the
set of all constants appearing in $\Pi$ and the \emph{Herbrand Base}
\HB{\Pi} be the set of all possible ground atoms which can be constructed
from the predicate symbols appearing in $\Pi$ with the constants of \HU{\Pi}.
Given a rule $r$, \GP{r} denotes the
set of rules obtained by applying all possible substitutions $\sigma$
from the variables in $r$ to elements of \HU{\Pi}. Similarly, given a
program $\Pi$, the {\em ground instantiation} \GP{\Pi} of $\Pi$
is the set \( \bigcup_{r \in \Pi} \GP{r} \).

For every program $\Pi$, its stable models are defined using its ground
instantiation \GP{\Pi} in two steps: First stable models of positive
programs are defined, then a reduction of general programs to positive
ones is given, which is used to define stable models of general
programs.

A set $L$ of ground literals is said to be {\em consistent} if, for every
literal $\ell \in L$, its negated literal $\naf \ell$ is not contained in
$L$.
Given a set of ground literals $L$, $\posOf{L} \subseteq L$ denotes
the set of positive literals in $L$.
An interpretation $I$ for $\Pi$ is a consistent set of ground literals
over atoms in $\HB{\Pi}$. %
A ground literal $\ell$ is {\em true} w.r.t.\ $I$ if $\ell\in I$;
$\ell$ is {\em false} w.r.t.\ $I$ if its negated literal is in $I$;
$\ell$ is {\em undefined} w.r.t.\ $I$ if it is neither true nor false w.r.t.\ $I$.
A constraint $c$ is said to be \textit{violated} by an interpretation $I$ if all literals in the body of $c$ are true.
An interpretation $I$ is {\em total} if, for each atom $a$ in $\HB{\Pi}$,
either $a$ or $\naf a$ is in $I$ (i.e., no atom in $\HB{\Pi}$ is undefined w.r.t.\ $I$).
Otherwise, it is \textit{partial}.
A total interpretation $M$ is a {\em model} for $\Pi$
if, for every $r \in \GP{\Pi}$, at
least one literal in the head of $r$ is true w.r.t.\ $M$ whenever all literals in the
body of $r$ are true w.r.t.\ $M$.
A model $X$ is a {\em stable model}
for a positive program $\Pi$ if any other model $Y$ of $\Pi$ is such that $\posOf{X} \subseteq \posOf{Y}$.

The {\em reduct} or {\em Gelfond-Lifschitz transform}
of a general ground program $\Pi$ w.r.t.\ an interpretation $X$ is the positive
ground program $\Pi^X$, obtained from $\Pi$ by (i) deleting all rules
$r \in \Pi$ whose negative body is false w.r.t.\ X and (ii)
deleting the negative body from the remaining rules.
A stable model of %
$\Pi$ is a model $X$ of $\Pi$ such
that $X$ is a stable model of $\GP{\Pi}^X$.
We denote by $SM(\Pi)$ the set of all stable models of $\Pi$,
and call $\Pi$ \textit{coherent} if $SM(\Pi) \neq \emptyset$, \textit{incoherent} otherwise.

\begin{example}\label{ex:grounding}
	Consider the following program $\Pi_1$:
	\begin{equation*}
	\begin{array}{lll}
	r_1: a(1) \leftarrow \naf b(1) &\qquad
	r_2: b(1) \leftarrow \naf a(1) & \qquad
	r_3: \leftarrow a(X), \ b(X) \\
	r_4: c(1) \leftarrow \naf d(1) & \qquad
	r_5: d(1) \leftarrow \naf c(1) & \qquad
	r_6: \leftarrow a(X), \ \naf b(X)
	\end{array}
	\end{equation*}
	The ground instantiation $\GP{\Pi_1}$ of the program $\Pi_1$ is the following program:
		\begin{equation*}
	\begin{array}{lll}
	g_1: a(1) \leftarrow \naf b(1) &\qquad
	g_2: b(1) \leftarrow \naf a(1) & \qquad
	g_3: \leftarrow a(1), \ b(1) \\
	g_4: c(1) \leftarrow \naf d(1) & \qquad
	g_5: d(1) \leftarrow \naf c(1) & \qquad
	g_6: \leftarrow a(1), \ \naf b(1)
	\end{array}
	\end{equation*}
	Note that %
  $M=\{\naf a(1), \ b(1), \ c(1), \ \naf d(1)\}$ is a model of $\GP{\Pi_1}$. Since $\GP{\Pi_1}^M$ comprises only the facts $b(1)$ and $c(1)$, and constraint $g_3$, $M$ is a stable model of $\Pi$.
$\hfill\lhd$
\end{example}

\paragraph{Support.}
Given a model $M$ for a ground program $\Pi$,
we say that a ground atom $a \in M$ is {\em supported}
with respect to $M$ if there exists a \emph{supporting} rule $r\in \Pi$
such that $a$ is the only true atom w.r.t. $M$ in the head of $r$, and all literals in the
body of $r$ are true w.r.t.\ $M$.
If $M$ is a stable model of a program $\Pi$, then all atoms in $M$ are supported.

\section{Solving Strategies}
\label{sec:solving}

\subsection{Classical Evaluation}\label{sec:solving:classical}

The standard solving approach for ASP is
instantiation followed by a procedure similar to CDCL for SAT
with extensions specific to ASP~\cite{DBLP:journals/aim/KaufmannLPS16}.
The basic algorithm
$\mi{ComputeStableModel}(\Pi)$
for finding a stable model of program $\Pi$
is shown in Algorithm~\ref{alg:mg}.
The Function~\ref{fn:propagatestd}
combines unit propagation (as in SAT) with
some additional ASP-specific propagations, which ensure
the model is stable (cf. \cite{DBLP:journals/aim/KaufmannLPS16}).

Given a partial interpretation $I$ consisting of literals,
and a set of rules $\Pi$,
\emph{unit propagation}
infers a literal $\ell$ to be true
if there is a rule $r \in \Pi$ such that
$r$ can be satisfied only by $I \cup \{\ell\}$.
Given the nogood representation
$C(r) = \{ \naf a_1, \ldots, \naf a_n, b_1, \ldots, b_j, \naf b_{j+1}, \ldots, \naf b_m \}$
of a rule $r$,
then the negation of a literal $\ell \in C(r)$
is unit propagated w.r.t.\ $I$ and rule $r$
iff $C(r) \setminus \{ \ell \} \subseteq I$.
To ensure that models are supported,
unit propagation is performed on the
Clark completion %
of $\Pi$ %
or alternatively a \emph{support propagator} is used~\cite{DBLP:conf/ijcai/AlvianoD16}.

\begin{example}\label{ex:prog}
Consider the ground program $\Pi_1$ from Example~\ref{ex:grounding}.
$\mi{ComputeStableModel}(\Pi_1)$
starts with $I = \emptyset$ and
does not propagate anything in line~\ref{ln:alg:propagate}.
$I$ is partial and consistent, so the algorithm continues
in line~\ref{ln:alg:partial}.
Assume no restart and no deletion is performed,
and assume $\mi{ChooseLiteral}$ returns $\{a(1)\}$, i.e., $I=\{a(1)\}$.
Next, $\mi{Propagate}(I)$ is called,
which yields $I = \{a(1),\  b(1),\  \naf b(1)\}$:
$\naf b(1)$ comes from unit propagation on $g_3$
and $b(1)$ from unit propagation on $g_6$.
Thus, $I$ is inconsistent and $I$ is analyzed to compute a reason explaining the conflict, i.e., $\mi{CreateConstraint}(I) = \{g_7\}$ with $g_7: \lar a(1)$.
Intuitively, the truth of $a(1)$ leads to an inconsistent interpretation, thus $a(1)$ must be false.
Then, the consistency of $I$ is restored (line~\ref{ln:alg:restore}), i.e., $I = \emptyset$, and $g_7$ is added to $\Pi_1$.
The algorithm again restarts at line~\ref{ln:alg:propagate}
with $I = \emptyset$
and propagates $I = \{\naf a(1),\ b(1)\}$,
where $\naf a(1)$ comes from unit propagation on $g_7$,
and $b$ from unit propagation on $g_2$.
$I$ is partial and consistent,
therefore lines~\ref{ln:alg:partial} and~\ref{ln:alg:choice} are executed.
Assume again that no restart and no constraint deletion
happens,
and that $\mi{ChooseLiteral}(I) = \{ c(1) \}$.
Therefore, the algorithm continues in line~\ref{ln:alg:propagate}
with $I = \{ \naf a(1), \ b(1), \ c(1) \}$.
Propagation yields $I = \{ \naf a(1),\  b(1),\ c(1),\ \naf d(1) \}$
because $\naf d(1)$ is support-propagated w.r.t.\ $g_4$ and $I$ (or unit-propagated w.r.t.\ the completion of $g_4$ and $I$).
$I$ is total and consistent, therefore the algorithm returns $I$ as the first stable model.
$\hfill\lhd$
\end{example}

For the performance of this search procedure,
several details are crucial:
learning effective constraints from inconsistencies as well as heuristics for restarting, constraint deletion,
and for choosing literals.

\newcommand{\cP}{\mathcal{P}}
\begin{algorithm}[t]
	\SetKwInOut{Input}{Input}
	\SetKwInOut{Output}{Output}
	\Input{A ground program $\cP$}
	\Output{A stable model for $\cP$ or $\bot$}
	\Begin{
		$I := \emptyset$\;
		$I := $ \textit{Propagate($I$)}\; \label{ln:alg:propagate}
		\uIf{$I$ is inconsistent}
		{
			$r$ := \textit{CreateConstraint($I$) \label{ln:alg:learning}\;
			$I := $ \textit{RestoreConsistency}($I$)\; \label{ln:alg:restore}
			\lIf{$I$ is consistent\label{ln:alg:addconstraint}}{$\cP := \cP \ \cup \ \{r\}$}\;}
			\lElse{\Return $\bot$\;}
		}
		\lElseIf{$I$ total}
		{
			{\Return $I$\;}
		}
		\Else{
			$I := $ \textit{RestartIfNeeded($I$)}; \qquad $\cP := $ \textit{DeleteConstraintsIfNeeded($\cP$)}\; \label{ln:alg:partial}
			$I := I \ \cup $ \textit{ChooseLiteral($I$)}\; \label{ln:alg:choice}
		}
		\textbf{goto}~\ref{ln:alg:propagate}\;
	}
	\caption{ComputeStableModel}\label{alg:mg}
\end{algorithm}

\begin{function}[t]
	$\mathcal{I} = I$\;
	\lFor{$\ell \in \mathcal{I}$} %
	{
		$\mathcal{I}$ := $\mathcal{I} \ \cup \ Propagation(\mathcal{I}, \ \ell)$ \; %
	}
	\Return $\mathcal{I}$\;
	\caption{Propagate($I$)}\label{fn:propagatestd}
\end{function}

\subsection{Lazy Constraints}\label{sec:lazy}
\begin{algorithm}[t]
\SetKwInOut{Input}{Input}
\SetKwInOut{Output}{Output}
\Input{A nonground program $\Pi$, a set of nonground constraints $C \subseteq \Pi$}
\Output{A stable model for $\Pi$ or $\bot$}
\Begin{
	$\mathcal{P}$ := $\GP{\Pi \setminus C}$\;
	$I$ := $ComputeStableModel(\mathcal{P})$\; \label{ln:lazy:search}
	\lIf{$I$ == $\bot$}{ \Return $\bot$ \label{ln:lazy:incoherent}\;}
	$\mathcal{C} = \{c \mid c \in \GP{C}, \ c $ is violated$\}$\label{ln:lazy:violating}\;
	\lIf {$\mathcal{C}$ == $\emptyset$}{\Return $I$\;}
	$\mathcal{P}$ := $\mathcal{P} \cup \mathcal{C}$ \label{ln:lazy:constraint}\;
	\textbf{goto}~\ref{ln:lazy:search}\;
}
\caption{LazyConstraintInstantiation}\label{alg:lazy}
\end{algorithm}

The algorithm presented in this section is reported as Algorithm~\ref{alg:lazy}.
The algorithm takes as input a program $\Pi$ and a set of constraints $C \subseteq \Pi$.
Then, the constraints in $C$ are removed from $\Pi$, obtaining the program $\mathcal{P}$.
A stable model of $\GP{\mathcal{P}}$ is searched (line~\ref{ln:lazy:search}).
Two cases are possible:
$(i)$ $\mathcal{P}$ is incoherent (line~\ref{ln:lazy:incoherent}). Thus, the original program $\Pi$ is also incoherent and the algorithm terminates returning $\bot$.
$(ii)$ $\mathcal{P}$ is coherent. Thus, a stable model, say $I$, is computed. In this case, a set of constraints $\mathcal{C} \in \GP{C}$ that are violated under the stable model $I$ are extracted (line \ref{ln:lazy:violating}) and added to $\mathcal{P}$ (line~\ref{ln:lazy:constraint}).
The process is repeated until either a stable model of $\mathcal{P}$ violating no constraints in $\GP{C}$ is found or $\mathcal{P}$ is incoherent.
Importantly, $\GP{C}$ is never represented explicitly in the implementation of line~5.

\begin{example}\label{ex:lazy}
Again consider program $\Pi_1$
from Example~\ref{ex:grounding}
and the set of constraints $C= \{r_3, \ r_6\}$.
The algorithm computes a stable model, say $I_1=\{a(1), \ \naf b(1), \ c(1), \ \naf d(1)\}$, of $\mathcal{P}_1 = \GP{\Pi_1\  \setminus \ C}$.
Thus, the ground instantiation $g_6$ of $r_6$ is violated under $I_1$ and therefore $g_6$ is added to $\mathcal{P}$.
Then, a stable model of $\mathcal{P}$ is computed, say $I_2=\{\naf a(1), \ b(1), \ c(1), \ \naf d(1)\}$.
At this point, $I_2$ violates no constraint in $\GP{C}$. Thus, the algorithm terminates returning $I_2$.
Note that all instantiations of constraint $r_3$ will be never violated since rules $r_1$ and $r_2$ enforce that exactly one of $a(1)$ and $b(1)$ can be true in a stable model. Thus, $r_3$ will never be instantiated by the algorithm.$\hfill \lhd$
\end{example}
An important feature of Algorithm~\ref{alg:lazy} is that it requires no modifications to the search procedure implemented by the underlying ASP solver.

\subsection{Constraints via Propagators}\label{sec:propagators}
In this section, constraints are replaced using the concept of \textit{propagator}, which can set truth values of atoms during the solving process,
based on truth values of other atoms.
An example of a propagator is the unit propagation, detailed in Section~\ref{sec:solving:classical}.
In contrast to the lazy instantiation of constraints that aims at adding violated constraints when a stable model candidate is found, propagators usually are used to evaluate the constraints during the computation of the stable model.
Given a program $\Pi$, traditional solvers usually apply propagators on the whole set of rules and constraints in $\GP{\Pi}$.
An alternative strategy is to consider a variant of the program, say $\mathcal{P} = \Pi \setminus C$, where $C$ is a set of constraints.
The solver is then executed on $\GP{\mathcal{P}}$ and a propagator is used to guarantee the coherence of partial interpretations with the constraints in $\GP{C}$.
Constraints in $C$ are not instantiated in practice but their inferences are simulated by an ad-hoc procedure
implemented for that purpose.
This approach requires a modification of the Propagation function in \ref{fn:propagatestd},
such that Propagation considers the additional set $C$ of constraints,
verifies which constraints would result in a propagation on the partial interpretation,
and propagate truth values due to inferences on $C$ in addition to unit propagation.

\begin{example}
Again consider program $\Pi_1$ from Example~\ref{ex:grounding} and the set of constraints $C= \{r_3, \ r_6\}$.
The idea is to execute Algorithm~\ref{alg:mg}
on $\GP{\mathcal{P}_1}$, where $\mathcal{P}_1 = \Pi_1 \setminus C$.
$\mi{ComputeStableModel}(\mathcal{P}_1)$
starts with $I = \emptyset$ and
does not propagate anything in line~\ref{ln:alg:propagate}.
$I$ is partial and consistent, so the algorithm continues
in line~\ref{ln:alg:partial}.
Assume no restart and no deletion is performed,
and assume $\mi{ChooseLiteral}$ returns $\{a(1)\}$, i.e., $I=\{a(1)\}$.
Next, $\mi{Propagate}(I, \ C)$ is called.
In this case, the propagation yields $I = \{a(1), \ b(1), \ \naf b(1)\}$, where $\naf b(1)$ comes from unit propagation on $g_1$, while $b(1)$ comes from unit propagation on the ground instantiation $g_6$ of the rule $r_6$.
Thus, $I$ is inconsistent and $I$ is analyzed to compute a reason that explains the conflict, i.e., $\mi{CreateConstraint}(I) = \{g_7\}$ with $g_7: \lar a(1)$.
Then, the algorithm continues as shown in Example~\ref{ex:prog}.
Note that, from this point of the computation, the ground instantiations of constraints $r_3$ and $r_6$ will never be violated again, since $g_7$ assure that $a(1)$ will be false in all partial interpretations under consideration. $\hfill \lhd$
\end{example}

We classify constraint propagators according to the priority given to them. In particular, they are considered \textit{eager} if propagation on non-ground constraints is executed as soon as possible, i.e., during unit propagation of already grounded constraints; moreover, they are called \textit{postponed} (or \textit{post}) if propagation on constraints is executed after all other (unit, support, etc.) propagations.

\section{Implementation and Experimental Analysis}\label{sec:experiments}

\subsection{Implementation}
\label{sec:implementation}
The lazy instantiation of constraints and the propagators have been implemented on top of the ASP solvers \wasp~\cite{DBLP:conf/lpnmr/AlvianoDLR15} and \clingo~\cite{DBLP:conf/iclp/GebserKKOSW16}.
The Python interface of \wasp~\cite{DBLP:conf/aiia/DodaroRS16} follows a synchronous message passing protocol implemented by means of method calls.
Basically, a Python program implements a predetermined set of methods that are later on called by \wasp whenever specific points of the computation are reached.
The methods may return some values that are then interpreted by \wasp.
For instance, when a literal is true the method \textit{onLiteralTrue} of the propagator is called, whose output is a list of literals to infer as true as a consequence (see~\cite{DBLP:conf/aiia/DodaroRS16} for further details).
\clingo~5~\cite{DBLP:conf/iclp/GebserKKOSW16}
provides a Python interface where a propagator class
with an interface similar to \wasp can be registered.

Two important differences exist between \wasp and \clingo.
Firstly \clingo provides only a post-propagator interface
and no possibility for realizing an eager propagator
(that runs before unit propagation is finished).
Secondly, \wasp first collects nogoods added in Python
and then internally applies them and handles conflicts,
while \clingo requires an explicit propagation call
after each added nogood.
If propagation returns a conflict
then no further nogoods can be added in \clingo,
even if further nogoods were detected.
After consulting the \clingo authors,
we implemented a queue for nogoods
and add them in subsequent propagations
if there is a conflict.
This yields higher performance than abandoning these nogoods.

\subsection{Description of Benchmarks}\label{sec:benchdescription}
In order to empirically compare the various strategies for avoiding the instantiation of constraints, we investigated several benchmarks of different nature, namely Stable Marriage, Packing, and Natural Language Understanding.
All benchmarks contain one or few constraints whose grounding can be problematic.

\paragraph{Stable Marriage.} The \emph{Stable Marriage} problem can be described as follows:
given $n$ men and $m$ women, where each person has a preference order over the opposite sex, marry them so that the marriage is stable.
In this case, the marriage is  said to be stable if there is no couple $(M, W)$ for which both partners would rather be married with each other than their current partner.
We considered the encoding used for the fourth ASP Competition.
For the lazy instantiation and for the ad-hoc propagators the following constraint has been removed from the encoding:
\begin{equation*}
\begin{array}{l}
\leftarrow match(M,W1), \ manAssignsScore(M,W,Smw), \ W1 \ne W, \\
\quad \quad manAssignsScore(M,W1,Smw1),  \ Smw > Smw1, \ match(M1,W),\\
\quad \quad  womanAssignsScore(W,M,Swm), \ womanAssignsScore(W,M1,Swm1), \ Swm \ge Swm1.
\end{array}
\end{equation*}
Intuitively, this constraint guarantees that the stability condition is not violated.

\paragraph{Packing.} The \emph{Packing Problem} is related to a class of problems in which one has to pack objects together in a given container. We consider the variant of the problem submitted to the ASP Competition 2011. In that case, the problem was the packing of squares of possibly different sizes in a rectangular space and without the possibility of performing rotations.
The encoding follows the typical guess-and-check structure, where positions of squares are guessed and some constraints check whether the guessed solution is a stable model.
We identified 2 expensive sets of constraints.
The first set comprises the following two constraints:
\begin{align*}
\leftarrow pos(I,X,Y), pos(I,X_1,Y_1), X_1 \neq X & \qquad
\leftarrow pos(I,X,Y), pos(I,X_1,Y_1), Y_1 \neq Y
\end{align*}
which enforce that a square is not assigned to different positions.
The second set comprises constraints forbidding the overlap of squares.
One of these constraints is reported in the following:
\begin{align*}
\leftarrow pos(I_1,X_1,Y_1), \ square(I_1,D_1), \ pos(I_2,X_2,Y_2), \ square(I_2,D_2), \phantom{we are free to stay herei}\\
I1 \neq I2, \ W1 = X1+D1, \
H1 = Y1+D1, \ X2 \geq X1,\ X2 < W1, \ Y2 \geq Y1, \ Y2 < H1.
\end{align*}
Other constraints are similar thus they are not reported.

\paragraph{Natural Language Understanding (NLU).} The \emph{NLU} benchmark is an application of ASP
in the area of Natural Language Understanding,
in particular the computation of optimal solutions
for First Order Horn Abduction problems
under the following cost functions:
cardinality minimality,
cohesion%
, and weighted abduction%
.
This problem and these objective functions have been described by
Sch\"uller~\shortcite{DBLP:journals/fuin/Schuller16}.
In this problem,
we aim to find a set of explanatory atoms
that makes a set of goal atoms true with respect to a
First Order Horn background theory.
We here consider the acyclic version of the problem
where backward reasoning over axioms
is guaranteed to introduce a finite set of
new terms.
A specific challenge in this problem is
that input terms and terms invented via backward chaining
can be equivalent to other terms,
i.e., the unique names assumption is partially not true.
Equivalence of terms must be handled explicitly in ASP,
which is done by guessing an equivalence relation.
This makes the instantiation of most instances
infeasible, as the number of invented terms becomes large,
due to the grounding blow-up caused by the following constraint:
\begin{align*}
&\leftarrow \mi{eq(A,B)},\, \mi{eq(B,C)},\, \naf \mi{eq(A,C)}.
\end{align*}

\subsection{Hardware and Software Settings}
The experiments were run on a Intel Xeon CPU X3430 2.4 GHz.
Time and memory were limited to 600 seconds and 4 GB, respectively.
In the following, \wasp-\textsc{lazy} refers to \wasp implementing lazy instantiation of constraints, while \wasp-\textsc{eager} and \wasp-\textsc{post} refer to \wasp implementing eager and postponed propagators, respectively.
All versions of \wasp use \gringo version 5.1.0 as grounder, whose grounding time is included in the execution time of \wasp.
Moreover, \clingo~\textsc{lazy} and \clingo~\textsc{post} refer to \clingo implementing lazy and postponed propagators, respectively. For the NLU benchmark, we always use unsat-core optimization.

\subsection{Discussion of Results}
\begin{table}[b!]
	\caption{Stable Marriage: Number of solved instances and average running time (in seconds).} \label{tab:stable}
	\centering
	\footnotesize
	\setlength{\tabcolsep}{0.2em}
	\begin{tabular}{rrrrrrrrrrrrrrrr}
		\toprule
		\textbf{Pref. (\%)}  & \multicolumn{2}{c}{\textbf{\wasp}}	& \multicolumn{2}{c}{\textbf{\wasp-\textsc{lazy}}} 	& \multicolumn{2}{c}{\textbf{\wasp-\textsc{eager}}} & \multicolumn{2}{c}{\textbf{\wasp-\textsc{post}}} & \multicolumn{2}{c}{\textbf{\clingo}}	& \multicolumn{2}{c}{\textbf{\clingo-\textsc{lazy}}}	& \multicolumn{2}{c}{\textbf{\clingo-\textsc{post}}}	&\\
		& \textbf{sol.} & \textbf{avg t}
		& \textbf{sol.} & \textbf{avg t} & \textbf{sol.} & \textbf{avg t}  & \textbf{sol.} & \textbf{avg t} & \textbf{sol.} & \textbf{avg t} & \textbf{sol.} & \textbf{avg t} & \textbf{sol.} & \textbf{avg t}\\
		    0	&	10	&	4.1	&	10	&	4.7	&	10	&	4.6	&	10	&	4.7 & 10    	&	10.6	&	10	&	4.2	&	10	&	4.2\\
			5	&	9	&	16.2	&	10	&	4.7	&	10	&	4.3	&	10	&	4.9 &   10	&	23.0	&	10	&	4.6	&	10	&	4.4\\
			10	&	10	&	19.2	&	10	&	4.7	&	10	&	4.3	&	10	&	4.6 &   10	&	34.6	&	10	&	6.4	&	10	&	8.2\\
			15	&	9	&	24.3	&	10	&	4.7	&	10	&	4.4	&	10	&	4.8 & 10	&	42.9	&	10	&	9.6	&	10	&	17.5\\
			20	&	8	&	35.2	&	10	&	4.8	&	10	&	4.6	&	10	&	5.2& 10	&	48.9	&	10	&	16.5	&	10	&	24.7\\
			25	&	10	&	34.8	&	10	&	4.8	&	10	&	5.4	&	10	&	6.0& 10	&	53.9	&	10	&	22.2	&	10	&	42.8\\
			30	&	6	&	97.0	&	10	&	5.0	&	10	&	7.7	&	10	&	7.6& 10	&	59.5	&	10	&	32.2	&	10	&	92.1\\
			35	&	10	&	42.1	&	10	&	5.0	&	10	&	8.2	&	10	&	10.0& 10	&	65.8	&	10	&	62.4	&	10	&	115.9\\
			40	&	9	&	51.3	&	10	&	5.2	&	10	&	7.6	&	10	&	9.2& 10	&	68.4	&	10	&	81.8	&	10	&	117.5\\
			45	&	10	&	113.4	&	10	&	5.4	&	10	&	10.8	&	10	&	12.0& 10	&	71.0	&	10	&	97.7	&	10	&	140.8\\
			50	&	6	&	74.6	&	10	&	5.1	&	10	&	22.4	&	10	&	20.3& 10	&	72.0	&	10	&	153.6	&	10	&	143.4\\
			55	&	9	&	44.5	&	8	&	5.9	&	10	&	39.4	&	10	&	23.6& 10	&	72.9	&	10	&	193.8	&	10	&	166.5\\
			60	&	9	&	70.9	&	10	&	7.7	&	10	&	23.8	&	10	&	25.0& 10	&	74.6	&	10	&	241.1	&	10	&	181.6\\
			65	&	7	&	99.3	&	10	&	11.4	&	10	&	64.7	&	10	&	54.2& 10	&	74.7	&	10	&	295.6	&	10	&	209.8\\
			70	&	9	&	89.3	&	5	&	25.5	&	10	&	121.8	&	10	&	101.8& 10	&	75.0	&	10	&	361.1	&	10	&	235.3\\
			75	&	8	&	77.0	&	0	&	-	&	10	&	184.0	&	10	&	146.7& 10	&	75.1	&	6	&	472.1	&	10	&	311.0\\
			80	&	7	&	85.5	&	0	&	-	&	10	&	248.6	&	8	&	274.7& 10	&	76.3	&	0	&	-	&	10	&	434.3\\
			85	&	4	&	259.5	&	0	&	-	&	10	&	232.3	&	1	&	337.2& 10	&	82.3	&	0	&	-	&	7	&	569.7\\
			90	&	9	&	79.2	&	0	&	-	&	5	&	449.4	&	0	&	-& 10	&	251.1	&	0	&	-	&	1	&	577.7\\
			95	&	10	&	46.3	&	0	&	-	&	0	&	-	&	0	&	-& 6	&	273.6	&	0	&	-	&	3	&	580.8\\
			100 &	8	&	67.6	& 1	&	81.2	&	10	&	133.3	&	10	&	153.6& 10	&	74.1	&	6	&	493.3	&	10	&	323.9\\
		\bottomrule
	\end{tabular}
\end{table}
\paragraph{Stable Marriage.}
Concerning Stable Marriage, we executed the 30 instances selected for the Fourth ASP Competition.
\clingo and \wasp executed on the full encoding are able to solve 29 out of the 30 instances with an average running time of 50 and 29 seconds, respectively.
On the same instances, ad-hoc propagators cannot reach the same performance.
Indeed, \wasp-\lazy and \wasp-\post perform the worst solving 0 and 5 instances, respectively, whereas \wasp-\eager is much better with 17 solved instances.
The same performance is obtained by \clingo-\lazy and \clingo-\post which can solve 0 and 17 instances in the allotted time, respectively.
The poor performance of the lazy instantiation can be explained by looking at the specific nature of the instances.
Indeed, each instance contains a randomly generated set of preferences of men for women (resp. women for men).
By looking at the instances we observed that each man (resp. woman)
has a clear, often total, preference order over each woman (resp. man).
This specific case represents a limitation for employing the lazy instantiation.
Indeed, \wasp and \clingo executed on the encoding without the stability constraint perform naive choices until a stable model candidate is found.
Then, each candidate contains several violations of the stability condition and many constraints are added.
However, those constraints are not helpful since they only invalidate the current stable model candidate.
In general, for instances where the program without the stability condition is under-constrained many stable model candidates need to be invalidated before an actual solution is found (intuitively, given a program $\Pi$ and a set of constraints $C \subseteq \Pi$,
$|SM(\Pi \setminus C)| \gg |SM(\Pi)|$).

In order to further analyze this behavior empirically, we have conducted an additional experiment on the same problem.
In particular, we randomly generated instances where each man (resp. woman) gives the same preference to each woman (resp. man), so basically the stability condition is never violated.
Then, we consider a percentage $k$ of preferences, i.e., each man (resp. woman) gives the same preference to all the women (resp. men) but to $k$\% of them a lower preference is given.
In this way, instances with small values of $k$ should be easily solved by lazy instantiation, whereas instances with high values of $k$ should be hard.
For each considered percentage $k$, we executed 10 randomly generated instances.
Results are reported in Table~\ref{tab:stable}, where the number of solved instances and the average running time are shown for each tested approach.
Concerning \wasp, as observed before, for instances where the value of $k$ is small (up to 50\%) the lazy approach can solve all the instances with an average running time of about 5 seconds.
On the other hand, for high values of $k$ the advantages of the lazy approach disappear, as observed for the competition instances.
Interestingly, the eager propagator obtained the best performance overall.
For the tested instances, it seems to benefit of a smaller program and generation of the inferences does not slow down the performance as observed for competition instances.
Concerning \clingo, the lazy approach is the best performing one for instances where the value of $k$ is up to 35\%.
As shown for \wasp, the performance of the lazy approach are worse for high values of $k$.

\paragraph{Packing.}
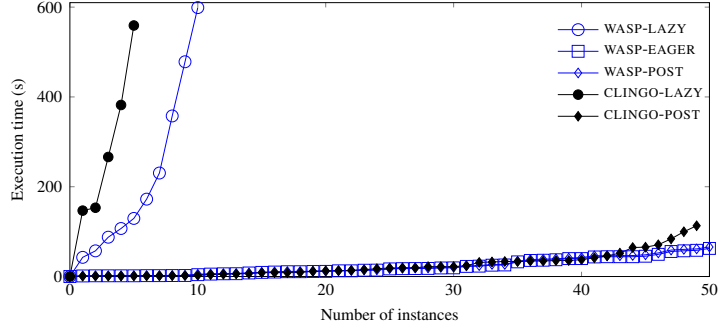
\begin{figure}[t]
	\figrule
	\begin{tikzpicture}[scale=0.9]
	\pgfkeys{%
		/pgf/number format/set thousands separator = {}}
	\begin{axis}[
	scale only axis
	, font=\scriptsize
	, x label style = {at={(axis description cs:0.5,0.04)}}
	, y label style = {at={(axis description cs:0.05,0.5)}}
	, xlabel={Number of instances}
	, ylabel={Execution time (s)}
	, xmin=0, xmax=50
	, ymin=0, ymax=610
	, legend style={at={(0.88,0.96)},anchor=north, draw=none,fill=none}
	, legend columns=1
	, width=0.7\textwidth
	, height=0.3\textwidth
	, ytick={0,200,400,600}
	, xtick={0,10,20,30,40,50}
	, major tick length=2pt
	]

	\addplot [mark size=2.5pt, color=blue, mark=o] [unbounded coords=jump] table[col sep=semicolon, y index=3] {./packing.csv};
	\addlegendentry{\wasp-\textsc{lazy}}

	\addplot [mark size=2.5pt, color=blue, mark=square] [unbounded coords=jump] table[col sep=semicolon, y index=1] {./packing.csv};
	\addlegendentry{\wasp-\textsc{eager}}

	\addplot [mark size=2pt, color=blue, mark=diamond] [unbounded coords=jump] table[col sep=semicolon, y index=2] {./packing.csv};
	\addlegendentry{\wasp-\textsc{post}}

	\addplot [mark size=2pt, color=black, mark=*] [unbounded coords=jump] table[col sep=semicolon, y index=5] {./packing.csv};
	\addlegendentry{\clingo-\textsc{lazy}}

	\addplot [mark size=2pt, color=black, mark=diamond*] [unbounded coords=jump] table[col sep=semicolon, y index=4] {./packing.csv};
	\addlegendentry{\clingo-\textsc{post}}
	\end{axis}

	\end{tikzpicture}
	\caption{Packing: Comparison of \textsc{lazy} and \textsc{propagators} approaches on 50 instances.}\label{fig:packing}
	\figrule
\end{figure}
Concerning Packing problem, we considered all 50 instances submitted to the Third ASP Competition.
Interestingly, when all constraints are considered none of the instances can be instantiated within the time limit.
Thus, \clingo and \wasp do not even start the computation of a stable model.
The grounding time substantially decreases when the two sets of expensive constraints described in Section~\ref{sec:benchdescription} are removed from the encoding.
Indeed, in this case, the grounding time on the tested instances is 5 seconds on average, with a peak of 16 seconds.
Results of the lazy constraint instantiation and of constraint propagators on the resulting program are reported in the cactus plot of Figure~\ref{fig:packing}.
The graph highlights that \wasp-\textsc{eager}, \wasp-\textsc{post}, and \clingo-\textsc{post} basically obtained the same performance.
Indeed, the first two solve all the tested instances with an average running time of~22 and~23 seconds, respectively, while \clingo-\textsc{post} solves 49 out of 50 instances with an average running time of 25 seconds.
Both \wasp-\textsc{post} and \clingo-\textsc{post} outperform their lazy counterparts.
Indeed, \wasp-\textsc{lazy} solves~10 instances, with an average running time of~226 seconds, while \clingo-\textsc{lazy} solves~5 instances, with an average running time of~301 seconds.
As already observed on the Stable Marriage instances, lazy instantiation cannot compete with constraint propagators.
In this experiment, we observed that \wasp and \clingo perform naive choices on the encoding without the expensive constraints, thus each candidate stable model contains several violations of constraints, leading to inefficient search in harder instances.

\paragraph{Natural Language Understanding (NLU).}
\begin{table}[b!]
	\caption{NLU Benchmark: Number of solved instances and average running time (in seconds).} \label{tab:nlu}
	\footnotesize
	\centering
	\setlength{\tabcolsep}{0.2em}
	\begin{tabular}{lrrrrrrrrrrrrrr}
		\toprule
		\textbf{Obj. Func.}  & \multicolumn{2}{c}{\textbf{\wasp}}	& \multicolumn{2}{c}{\textbf{\wasp-\textsc{lazy}}} 	& \multicolumn{2}{c}{\textbf{\wasp-\textsc{eager}}} & \multicolumn{2}{c}{\textbf{\wasp-\textsc{post}}} & \multicolumn{2}{c}{\textbf{\clingo}} & \multicolumn{2}{c}{\textbf{\clingo-\textsc{lazy}}} & \multicolumn{2}{c}{\textbf{\clingo-\textsc{post}}}\\
		& \textbf{sol.} & \textbf{avg t} & \textbf{sol.} & \textbf{avg t} & \textbf{sol.} & \textbf{avg t}  & \textbf{sol.} & \textbf{avg t} & \textbf{sol.} & \textbf{avg t}  & \textbf{sol.} & \textbf{avg t}  & \textbf{sol.} & \textbf{avg t} \\
		Card.	&	43	&	39.7	&	50	&	2.3	&	50	&	4.3	&	50	&	3.3 & 41& 30.7  & 50 & 4.5 & 50 & 1.5\\
		Coh. 	&	43	&	40.1	&	50	&	18.5	&	50	&	8.8	&	50	&	6.3 & 41 & 30.7	& 49 & 24.6 & 49 & 15.8\\
		W. Abd.	&	43	&	49.3	&	50	&	26.6	&	49	&	66.1	&	50	&	62.6 & 41 & 33.9 & 48 & 31.9 & 50 & 24.0\\
		\bottomrule
	\end{tabular}
\end{table}
Concerning NLU, we considered all 50 instances and all three objective functions used in \cite{DBLP:journals/fuin/Schuller16}.
Results are reported in Table~\ref{tab:nlu}.
As a general observation, all the tested instances are solved by \wasp-\textsc{lazy} and \wasp-\textsc{post}, no matter the objective function.
Moreover, \wasp-\textsc{lazy} is on average faster than all other alternatives for both the objective functions cardinality and weighted abduction.
The good performance of lazy instantiation is related to the small number of failing stable model checks performed.
Indeed, only 2, 16, and 64 invalidations are on average required for cardinality, coherence, and weighted abduction, respectively.
The number of propagation calls is much higher for
\wasp-\textsc{eager} than for \wasp-\textsc{post}
(approximately \wasp-\textsc{eager} performs 3 times more propagation calls than \wasp-\textsc{post}).
However, the number of propagated literals that are not
immediately rolled back because of a conflict
is very similar,
hence it is clear that \wasp-\textsc{eager} performs
a lot of unnecessary propagations in this benchmark
and \wasp-\textsc{post} should be preferred.
Note that this is not generally the case for other benchmarks.
Concerning \clingo,
45, 248, and 321, stable model candidates are invalidated with \clingo-\lazy, respectively,
and a similar amount (26, 589, and 700, respectively) with \clingo-\post.
This shows that \clingo tends to produces more stable models
that violate lazy constraints.
These violations are detected earlier with \clingo-\post,
therefore it outperforms \clingo-\lazy in all objectives.
None of the \clingo propagators is able to solve all instances with all objectives,
whereas \wasp-\post solves all of them within 600~s.
In particular for objective functions cardinality and coherence,
\wasp is always slightly faster and uses slightly more memory than \clingo.
For weighted abduction, \clingo-\post is most efficient with \wasp-\lazy in second place.
Nevertheless, using \clingo or \wasp with a \lazy or \post propagator will always be an advantage
over using the pure ASP encoding where the constraints are instantiated prior to solving.
Hence the choice of the method for instantiating constraints
is more important than the choice of the solver.

\paragraph{Discussion.}\label{par:discussion}
We empirically investigated whether lazy instantiation or propagators can be a valid option for enhancing the traditional ``ground+solve'' approach.
When the full grounding is infeasible, then both lazy instantiation and propagators can overcome this limitation, even though they exhibit different behaviors depending on the features of the problem and of the instances.
This is particularly evident in Packing, where no instance can be grounded within the time limit.
Since propagators are activated during the search, while lazy instantiation intervenes only when a total interpretation is computed, propagators are preferable when the problematic constraint is important to lead the search toward a solution (as overlap constraints in Packing).
On the other hand, a high number of unnecessary propagations can make propagators inefficient and even slower than the lazy approach. In these cases, we observed that post propagators are better than eager propagators as remarked by the results on the objective function `weighted abduction' in the NLU benchmark.
The experiment on Stable Marriage highlights that lazy instantiation is effective when few constraints are instantiated during the search.
This is the case when: (i) it is very likely that a stable model of the simplified (i.e., without problematic constraints) input program also satisfies the lazy constraints; or (ii) the solver heuristics is such that one of the first candidate total interpretations also satisfies the lazy constraints.
This is also confirmed in the NLU benchmark where the instances often have the above characteristics, and the propagator is better only when the constraints generated by the lazy approach do not fit the working memory.
Moreover, from case (ii), we conjecture that the lazy approach can be effective in combination with domain-specific heuristics~\cite{DBLP:conf/aaai/GebserKROSW13,DBLP:journals/tplp/DodaroGLMRS16}.

\begin{figure}
\figrule
\begin{subfigure}{.5\textwidth}
	\begin{tikzpicture}
	\pgfkeys{%
		/pgf/number format/set thousands separator = {}}
	\begin{axis}[
	scale only axis
	, font=\scriptsize
	, x label style = {at={(axis description cs:0.5,0.04)}}
	, y label style = {at={(axis description cs:0.05,0.5)}}
	, xlabel={Number of clauses / number of variables}
	, ylabel={Execution time (s)}
	, xmin=2.5, xmax=5.5
	, ymin=0.04, ymax=40
	, log basis y=10, ytickten={-1,0,1}
	, ymode=log
	, legend style={at={(0.25,0.85)},anchor=north, draw=none,fill=none, font=\tiny}
	, legend columns=1
	, width=0.65\textwidth
	, height=0.3\textwidth
	, xtick={3,3.5,4,4.5,5}
	, major tick length=2pt
	]

	\addplot [mark size=1.5pt, color=blue, mark=x] [unbounded coords=jump] table[col sep=semicolon, y index=2] {./3-sat-220.csv};
	\addlegendentry{\wasp-\textsc{lazy}}

	\addplot [mark size=1.5pt, color=red, mark=o] [unbounded coords=jump] table[col sep=semicolon, y index=1] {./3-sat-220.csv};
	\addlegendentry{\wasp-\textsc{eager}}

	\draw[dotted] (axis cs: 4.26, 0.04) -- (axis cs: 4.26, 40);

	\end{axis}

	\begin{axis}[
	scale only axis
	, axis y line*=right
	, font=\scriptsize
	, x label style = {at={(axis description cs:0.5,0.04)}}
	, y label style = {at={(axis description cs:0.05,0.5)}}
	, xlabel={Number of clauses / number of variables}
	, ylabel={Frequency of UNSAT}
	, ylabel near ticks, yticklabel pos=right
	, xmin=2.5, xmax=5.5
	, ymin=-0.05, ymax=1.05
	, legend style={at={(0.25,0.98)},anchor=north, draw=none,fill=none, font=\tiny}
	, legend columns=1
	, width=0.65\textwidth
	, ytick={0,0.5,1}
	, height=0.3\textwidth
	, xtick={3,3.5,4,4.5,5}
	, major tick length=2pt
	]

	\addplot [dashed, mark size=1pt, color=black] [unbounded coords=jump] table[col sep=semicolon, y index=1] {./phaseTransition.csv};
	\addlegendentry{\textsc{unsat freq.}}
	\end{axis}

	\end{tikzpicture}
  \caption{Results with 220 variables}\label{fig:3-sat-220}
\end{subfigure}%
\begin{subfigure}{.5\textwidth}
	\begin{tikzpicture}
	\pgfkeys{%
		/pgf/number format/set thousands separator = {}}
	\begin{axis}[
	scale only axis
	, axis y line*=left
	, font=\scriptsize
	, x label style = {at={(axis description cs:0.5,0.04)}}
	, y label style = {at={(axis description cs:0.05,0.5)}}
	, xlabel={Number of clauses / number of variables}
	, ylabel={Execution time (s)}
	, xmin=2.5, xmax=5.5
	, ymin=0.04, ymax=4000
	, ymode=log
	, legend style={at={(0.25,0.85)},anchor=north, draw=none,fill=none, font=\tiny}
	, legend columns=1
	, width=0.65\textwidth
	, height=0.3\textwidth
	, xtick={3,3.5,4,4.5,5}
	, major tick length=2pt
	]

	\addplot [mark size=1.5pt, color=blue, mark=x] [unbounded coords=jump] table[col sep=semicolon, y index=2] {./3-sat-280.csv};
	\addlegendentry{\wasp-\textsc{lazy}}

	\addplot [mark size=1.5pt, color=red, mark=o] [unbounded coords=jump] table[col sep=semicolon, y index=1] {./3-sat-280.csv};
	\addlegendentry{\wasp-\textsc{eager}}

	\draw[dotted] (axis cs: 4.28, 0.04) -- (axis cs: 4.28, 4000);

	\end{axis}
	\begin{axis}[
	scale only axis
	, axis y line*=right
	, font=\scriptsize
	, x label style = {at={(axis description cs:0.5,0.04)}}
	, y label style = {at={(axis description cs:0.05,0.5)}}
	, xlabel={Number of clauses / number of variables}
	, ylabel={Frequency of UNSAT}
	, ylabel near ticks, yticklabel pos=right
	, xmin=2.5, xmax=5.5
	, ymin=-0.05, ymax=1.05
	, legend style={at={(0.25,0.98)},anchor=north, draw=none,fill=none, font=\tiny}
	, legend columns=1
	, width=0.65\textwidth
	, ytick={0,0.5,1}
	, height=0.3\textwidth
	, xtick={3,3.5,4,4.5,5}
	, major tick length=2pt
	]

	\addplot [dashed, mark size=1pt, color=black] [unbounded coords=jump] table[col sep=semicolon, y index=2] {./phaseTransition.csv};
	\addlegendentry{\textsc{unsat freq.}}

	\end{axis}

	\end{tikzpicture}
  \caption{Results with 280 variables}\label{fig:3-sat-280}
\end{subfigure}
\caption{3-SAT experiments. Red and blue lines correspond to eager propagators and lazy instantiation respectively. The dashed black line represents the percentage of UNSAT instances, while the vertical dotted line evidences the phase transition point (frequency is about $0.5$ at $R=4.26$).}%
\label{fig:threesat}
\figrule
\end{figure}
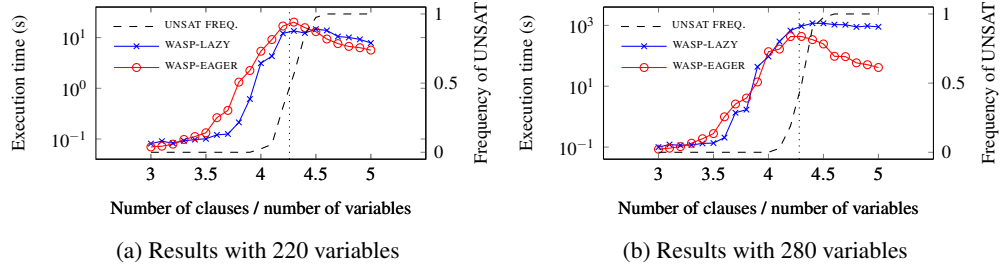

Finally, we conducted an additional experiment, where we do not oppose our approaches with the ground+solve one as in the previous cases, but it only aims at comparing the lazy propagation versus propagators in a controlled setting.
In particular, we considered a synthetic benchmark based on the well-known 3-SAT problem that is interesting for our study since it allows us to control both the hardness of the instances and the probability that an interpretation satisfies the constraint.
Indeed, we generated the instances uniformly at random in a range centered on the phase transition~\cite{DBLP:series/faia/Achlioptas09}. %
We used a straightforward ASP encoding where we guess an interpretation and we check by a single (non-ground) constraint whether this satisfies all clauses.
The results are summarized in Figure \ref{fig:threesat} where we present two representative runs on formulas with 220 and 280 Boolean variables, respectively.
Since eager and post propagators behave very similarly we only show comparisons between eager propagator and a lazy instantiation.

Expectedly, execution times follow the easy-hard-easy pattern~\cite{DBLP:series/faia/Achlioptas09}, centered on the phase transition, while varying the ratio $R$ of clauses over variables. Initially, the problem is very easy and both approaches are equally fast. Then there is an interval in which the lazy approach is preferable, and finally the eager approach becomes definitely better than the lazy.
Note that, on formulas with 220 variables (see Figure~\ref{fig:3-sat-220}) the lazy approach is preferable also on the hardest instances, instead with 280 variables (see Figure~\ref{fig:3-sat-280}) the eager approach becomes more convenient before the phase transition.
To explain this phenomenon we observe that the lazy approach can be exemplified by assuming that the solver freely guesses a model and then the lazy instantiator checks it, until every clause is satisfied by an assignment or no model can be found.
The probability that a random model satisfies all clauses is $(\frac{7}{8})^k$ where $k$ is the number of clauses, thus fewer tries are needed on average to converge to a solution if the formula has fewer clauses.
This intuitively explains why, as the number of variables increases, the eager approach becomes more convenient at smaller and smaller values of $R$.
It is worth pointing out that this simplified model does not fully capture the behavior of lazy instantiation that is more efficient in practice, since the implementation learns from previous failures (by instantiating violated constraints).

\portfoliotable{b!}

\subsection{On the applicability of techniques for automatic algorithm selection}\label{sec:portfolio}
The analysis conducted up to now shows that there is not always a clear winner among the strategies for realizing constraints,
since the best solving method depends on characteristics of the encoding and the instance at hand.
In similar scenarios, portfolio approaches
which automatically choose one out of a set of possible methods have proven to be very effective in increasing system performance, since they combine the strengths of the available methods.
Therefore, we investigated whether algorithm selection techniques can improve performance in our context.%

We apply basic algorithm selection based on classification
with machine learning:
we extract some natural features from each instance,
and train a C4.5 \cite{Quinlan1993c45} classifier
to predict the best solving method
(i.e., the one that required least amount of time) among all the available ones (including the plain solver).
We limit our analysis to Stable Marriage and NLU,
because in these domains none of the available methods is clearly superior.
As features for stable marriage we used the number of persons and the percentage or preferences, %
for NLU we used the number of facts and the number of distinct constants and (instance-specific) predicates.
We create portfolios for both \wasp-based and \clingo-based implementations.

Table \ref{tab:portfolio} shows the results
of our evaluation using 10-fold cross-validation
(i.e., we split the set of instances into 10 partitions
and use each partition as test set while training on the remaining partitions).
For each problem we report (weighted average) precision, recall, and f-measure of the prediction, as well as the average performance gain of the portfolio (i.e., by gain we mean the difference in percentage between the sum of the execution times measured for the portfolio and for its best method).
We observe that the classifier is able to choose the best algorithm in many cases, and the choice is almost ideal in NLU (f-measure of 0.9 for \wasp and 0.84 for \clasp). The portfolios are always faster (in terms of execution times) than the corresponding best method for the respective problem.
The performance gain peaks to 38\%  for the \wasp-based, and is less pronounced for the \clingo-based (peak at 13.6\%).
This is expected since \clingo features a basic solver that is more competitive with propagator-based solutions in these domains.

Summarizing, these results confirm that already the application of basic portfolio techniques is a viable option for improving the performance when propagators are available.

\section{Related Work}

The grounding bottleneck in ASP has been subject of various studies.
The most prominent grounding-less approach that works on plain ASP is \textit{lazy grounding}, which was implemented by \textsc{asperix}~\cite{DBLP:conf/lpnmr/LefevreN09a}, \textsc{gasp}~\cite{DBLP:journals/fuin/PaluDPR09}, and \textsc{omiga}~\cite{DBLP:conf/jelia/Dao-TranEFWW12}.
Differently from our approach that is focused on constraints,
these solvers perform lazy instantiation for \emph{all} the rules of a program,
and do not perform (conflict) clause learning.
Weinzierl~\shortcite{Weinzierl2017} recently investigated learning of non-ground clauses.

Lazy instantiation of constraints was topic of
several works on \emph{integrating ASP with other formalisms}.
These include CASP~\cite{DBLP:conf/asp/BaseliceBG05,DBLP:journals/tplp/OstrowskiS12,DBLP:journals/corr/BalducciniL17},
ASPMT~\cite{DBLP:conf/iclp/SusmanL16}, BFASP~\cite{DBLP:journals/tplp/AzizCS13}, and
HEX \cite{DBLP:journals/tplp/EiterFIKRS16}.
Differently from our approach, these approaches are based on syntactic extensions
that enable the combination of ASP solvers with solvers for external theories.
HEX facilitates the integration
of generic computation oracles as literals in ASP rule bodies,
and allows these computations not only to return true or false,
but also to inject constraints into the search.
This gave rise to the `on-demand constraint' usage pattern
of external atoms \cite{DBLP:conf/birthday/EiterRS16}
which roughly corresponds with the \lazy propagators in this work.
HEX also permits a declarative specification of
properties of external computations \cite{DBLP:journals/tplp/Redl16},
e.g., antimonotonicity with respect to some part of the model.
Such specifications automatically generate additional
lazy constraints. %
Integration of ASP with continuous motion planning in robotics,
based on HEX, was investigated in \cite{DBLP:journals/aicom/ErdemPS16}:
adding motion constraints in a \post propagator
was found to be
significantly faster than checking only complete stable model candidates (\lazy).
For integrating \emph{CModels with BProlog}
\cite{DBLP:journals/tplp/BalducciniL13}
it was shown that using BProlog similar as a \post propagator
(clearbox)
performs better than using it as a \lazy propagator
(black-box).

De~Cat et al.~\shortcite{DBLP:journals/jair/CatDBS15}
provide a theory and implementation
for \emph{lazy model expansion}
within the FO(ID) formalism
which is based on \emph{justifications}
that prevent instantiation of certain constraints
under assumptions.
These assumptions are relative to a model candidate
and can be revised from encountered conflicts,
leading to a partially lazy instantiation
of these constraints.

We finally observe that lazy constraints can be seen as a simplified form of lazy clause generation that was originally introduced in Constraint Programming~\cite{DBLP:conf/cp/FeydyS09}.

\section{Conclusion}\label{sec:conclusion}
In this paper, we compared several solutions for addressing the problem of the grounding bottleneck focusing on the practically-relevant case of problematic constraints without resorting to any language extension.
The considered approach can be seen as a natural extension of the ``ground+solve'' paradigm, adopted by  state of the art ASP systems, where some constraints are replaced either by lazy instantiators or propagators. The solutions fit CDCL-based solving strategies, and can be implemented using APIs provided by state of the art solvers.

Experiments conducted on both real-world and synthetic benchmarks clearly outline that all the approaches can solve instances that are out of reach of state of the art solvers because of the grounding blowup.
Lazy instantiation is the easiest to implement, and it is the best choice when the problematic constraints have a high probability to be satisfied. Otherwise, eager and post propagators perform better, with the latter being slightly more efficient when the constraint is activated more often during propagation.
Our empirical analysis shows that there is not always a clear winner for a given problem, thus we investigated the applicability of algorithm selection techniques.
We observed that a basic portfolio can improve on the best strategy also on these cases.
As far as future work is concerned, we will study what are the conditions under which an entire subprogram (and not just some constraints) can be replaced by a propagator. Another line of research might be to investigate the impact of applying rule decomposition techniques before handling the constraints~\cite{DBLP:journals/tplp/BichlerMW16}.

\section*{Acknowledgments}
The paper has been partially supported by the Italian Ministry for Economic Development (MISE) under project ``PIUCultura -- Paradigmi Innovativi per l'Utilizzo della Cultura'' (n. F/020016/01-02/X27),
under project ``Smarter Solutions in the Big Data World (S2BDW)'' (n. F/050389/01-03/X32) funded
within the call ``HORIZON2020'' PON I\&C 2014-2020,
and by the Scientific and Technological Research Council of Turkey
(TUBITAK) Grant 114E777.

\ifinlinerefs
\input{references.sty}
\else
\bibliographystyle{acmtrans}
\bibliography{references}
\fi

\label{lastpage}
\end{document}